\documentclass[11pt]{article}
\usepackage[preprint]{acl}
\usepackage{times}
\usepackage[T1]{fontenc}
\usepackage[utf8]{inputenc}
\usepackage{microtype}
\usepackage{inconsolata}
\usepackage{graphicx}
\usepackage{booktabs,multirow,array}
\usepackage{amsmath,amssymb}
\usepackage{xcolor}
\usepackage{url}
\usepackage{fontawesome5}
\newcommand{\bfcl}{BFCL\,v4}
\newcommand{\methodname}{StateAlign}
\newcommand{\hficon}{\raisebox{-0.18em}{\includegraphics[height=1.05em]{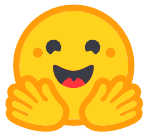}}}
\title{\methodname{}: Internalizing State-Aware Decision-Making for Reliable Multi-Turn Tool Agents via Reinforcement Learning}
\author{
  \textbf{Zhexi Lu\textsuperscript{1}},
  \textbf{Subhajit Chaudhury\textsuperscript{2}},
  \textbf{Tejaswini Pedapati\textsuperscript{2}}, \\
  \textbf{Keerthiram Murugesan\textsuperscript{2}},
  \textbf{Lei Yu\textsuperscript{1}} \\
  \textsuperscript{1}Rensselaer Polytechnic Institute,
  \textsuperscript{2}IBM Research \\[1ex]
  \normalsize
  \faGithub\ \textbf{Code:} \href{https://github.com/ZhexiLu/StateAlign}{\texttt{github.com/ZhexiLu/StateAlign}} \\
  \hficon\ \textbf{Model:} \href{https://huggingface.co/ZhexiLu/StateAlign-8B}{\texttt{huggingface.co/ZhexiLu/StateAlign-8B}}
}
\begin{document}
\maketitle
\begin{abstract}
Reinforcement learning has improved the tool-use capabilities of large language models. However, better task performance can coincide with increased tool hallucination. Reliable tool use requires decisions to follow the current task state: available information, tool capabilities, and task progress. Our analysis shows that agents sometimes execute when they should ask for information, retrieve it, or decline an infeasible request, even when reasoning identifies missing information. We introduce \methodname{}, combining task-state construction with state-contrastive reinforcement learning. The data cover five task conditions and pair states with different action requirements. Outcome rewards guide task completion; token-level contrasts compare the same generated reasoning and actions under paired states to train outputs that fit the current state. Experiments across five base models from three families show improvements on \bfcl{} multi-turn, ToolSandbox, and When2Call. Ablations show that state construction improves performance and reliability under different reward designs. The complete method further reduces tool hallucination across three Qwen3 sizes. On Qwen3-8B, \methodname{} improves \bfcl{} multi-turn accuracy from 45.5\% to 55.2\% and reduces no-tool hallucination on SimpleToolHalluBench from 62.2\% under outcome-only training to 56.1\%. Behavioral analyses show more consistent decisions, more complete multi-call execution, and stronger recovery after tool errors.
\end{abstract} 

\begin{figure}[t]
\centering
\includegraphics[width=\columnwidth]{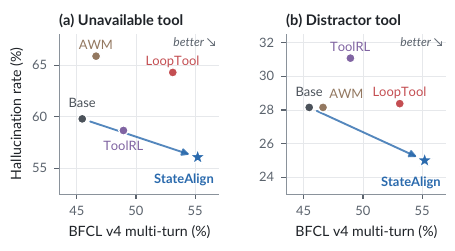}
\caption{Comparison of tool-use performance and reliability on Qwen3-8B. We plot \bfcl{} multi-turn accuracy against SimpleToolHalluBench hallucination rates with (a) no available tool and (b) only an irrelevant tool. Arrows connect the base model to \methodname{}. \methodname{} achieves the highest multi-turn accuracy and the lowest hallucination rates among the models shown.}
\label{fig:teaser}
\end{figure}
\section{Introduction}
\label{sec:introduction}

Tool calling allows large language models (LLMs) to interact with external tools and APIs \citep{schick2023toolformer}. These tools enable models to retrieve information, perform calculations, and execute operations requested by users. By incorporating tool results into subsequent reasoning, models can gather information and complete tasks over multiple steps \citep{yao2023react}. At each step, an agent must select an appropriate function, supply its arguments, and use the returned result to determine how to proceed.

To develop these capabilities, existing work constructs tool-use demonstrations for supervised fine-tuning (SFT). Datasets such as ToolBench, APIGen, and ToolACE \citep{qin2024toolllm,liu2024apigen,liu2025toolace} pair user requests and tool descriptions with target function calls and their arguments, teaching models which tools to select and how to invoke them. APIGen-MT, ToolACE-MT, and ToolMind \citep{prabhakar2025apigenmt,zeng2026toolacemt,yang2025toolmind} extend this supervision to multi-turn conversations with user messages, tool calls, and tool responses. However, SFT trains the model to imitate reference actions from fixed demonstrations. During interaction, errors in the model's own calls can lead to histories not covered by these demonstrations, where it may struggle to recover and complete the task.

Reinforcement learning (RL) instead trains models with reward feedback on their own outputs. Recent work improves tool use through reward design \citep{qian2025toolrl,zhang2026tooln1}, iterative data refinement \citep{zhang2026looptool}, and interactive training with code execution or synthesized tool environments \citep{feng2026retool,wang2026awm}. However, higher task performance does not necessarily mean more reliable tool use. Prior work has documented hallucinations in tool selection and use \citep{xu2025relign}, and stronger reasoning can further increase tool hallucination \citep{yin2026reasoning}. As shown in Figure~\ref{fig:teaser}, some RL-trained agents achieve higher \bfcl{} multi-turn accuracy than the base model but also exhibit higher tool hallucination rates. To understand this problem, we analyze whether agents choose appropriate actions given the information and tools available at each step. Our analysis in Section~\ref{sec:task-state-decisions} shows that agents often proceed with execution when they should ask for missing information, retrieve it through a tool, or decline an infeasible request. A model can even identify missing information in its reasoning but still generate a tool call with fabricated arguments. These findings motivate us to train agents to decide whether and how to act based on the current task state.

We introduce \methodname{}, a reinforcement learning framework for state-aware, reliable tool use. Starting from tool-use trajectories, we construct task states with different information availability, tool capabilities, and task progress. The model is trained on these states with outcome rewards based on task completion. We further design a token-level state-contrastive reward that compares the same generated reasoning and actions under paired states to encourage outputs consistent with the current task state and reduce tool hallucination. Our contributions are:

\begin{figure}[t]
\centering
\includegraphics[width=\columnwidth]{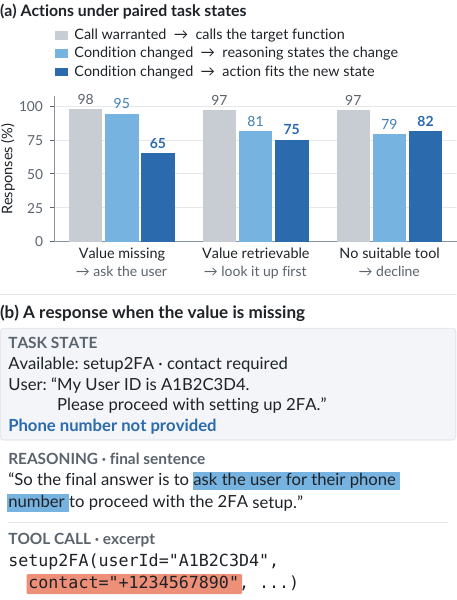}
\caption{Task-state diagnosis on Qwen3-8B. \textbf{(a)} Each group pairs a state in which calling the target function is warranted with one that differs in a single condition; bars give the share of responses. \textbf{(b)} Response excerpts when a required phone number is missing. Blue shading marks the decision to ask; red shading marks the unprovided value in the emitted call.}
\label{fig:task_state_diagnosis}
\end{figure}

\begin{figure*}[t]
\centering
\includegraphics[width=\textwidth]{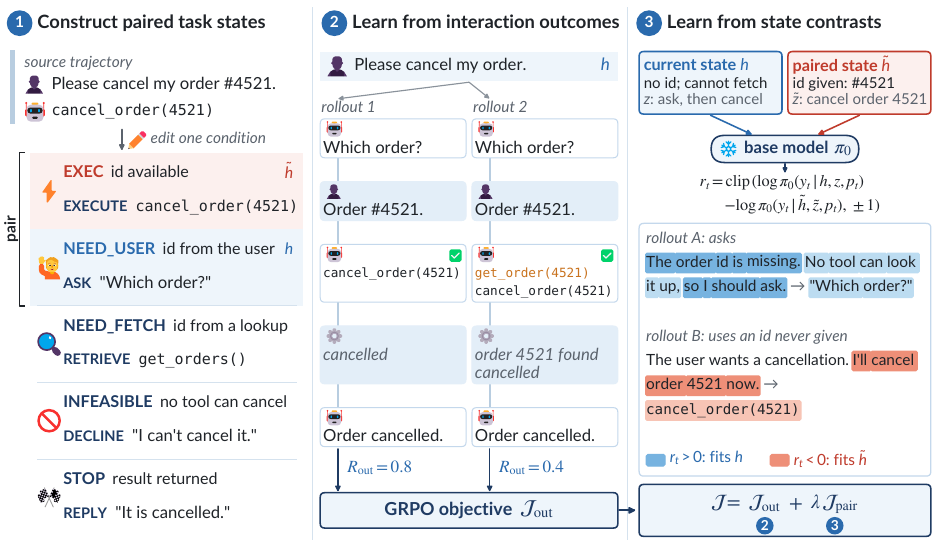}
\caption{Overview of \methodname{} on an example task. (1) Editing one condition of a step from a source trajectory yields five task states with different required actions; each state keeps a reference path, and states are paired with \textsc{Exec}. (2) The policy acts in a constructed state; shaded cards are user messages and tool results, taken from the source trajectory where available. Both rollouts ask for the order ID and then execute; the second also makes an unnecessary call in the same step, which halves its reward. Their outcome rewards give group-relative advantages in the GRPO objective $\mathcal J_{\mathrm{out}}$. (3) The base model $\pi_0$, kept fixed during training, scores the tokens of two further rollouts under the current state and the paired state, each with its reference path; the clipped difference $r_t$ weights the policy's token log-likelihoods in $\mathcal J_{\mathrm{pair}}$, defined in Eq.~\ref{eq:pair-objective}.}
\label{fig:overview}
\end{figure*}

\begin{itemize}
    \item We introduce a task-state construction method that converts tool-use trajectories into training examples covering diverse task conditions, with explicit pairs requiring different actions. Compared with the original training data, the constructed states improve task performance and reliability under different reward designs.
    \item We propose state-contrastive RL that combines interaction outcome rewards with token-level supervision from paired task states. On the same constructed data, \methodname{} reduces tool hallucination across three Qwen3 model sizes compared with training on task outcome rewards alone.
    \item Experiments across five base models show improved performance on \bfcl{} multi-turn, ToolSandbox, and When2Call. Behavioral analyses show more consistent decisions across task states, more complete multi-call execution, and stronger recovery after \mbox{tool errors}.
\end{itemize}

\section{Task States and Agent Decisions}
\label{sec:task-state-decisions}

\paragraph{Task State.}
A tool agent receives a context $h=(\mathcal T,H)$ containing tool definitions $\mathcal T$ and interaction history $H$. For a user goal, we define the \emph{task state} as the task-relevant conditions specified by the tools and evidenced by the history that constrain the next action. These include (i)~\emph{information availability}: what information needed for the task is available, what is missing, and how it can be obtained; (ii)~\emph{tool capabilities}: which operations the tools support; and (iii)~\emph{task progress}: what is complete and what remains. User replies, tool results, and tool-availability changes can update these conditions and require different actions. Prior work models decisions over interaction histories \citep{zhou2024archer} and combines dialogue-state tracking with tool-use training \citep{acikgoz2025coalm}.

\paragraph{Decisions across Task States.}
We compare Qwen3-8B's responses to paired task states following the protocol in Appendix~\ref{app:diagnostics}. The states differ in one relevant condition: one allows execution, and the other requires clarification, retrieval, or declining an infeasible request. Figure~\ref{fig:task_state_diagnosis}a shows that target-function selection reaches 97--98\% when execution is appropriate, but choosing the required action under the contrasting conditions is less reliable. In the missing-information condition, 95\% of responses identify the missing value in reasoning, but only 65\% ask the user for it. Figure~\ref{fig:task_state_diagnosis}b shows the model ending its reasoning with a decision to ask for the user's phone number, then calling the tool with a number absent from the context. The final call contradicts both the available information and the decision expressed in reasoning.

The training problem is therefore to make reasoning and actions follow the conditions of the current task. We construct training states with different action requirements and pair states that differ in a relevant condition. For example, reasoning that assumes a required value is available may fit a state that permits execution but conflict with a state requiring clarification. The same generated content can then be evaluated under both states. \methodname{} combines this supervision with feedback on interaction outcomes to train reasoning and actions that fit the current state.

\section{\methodname{}: State-Aware Reinforcement Learning}
\label{sec:method}

We illustrate \methodname{} in Figure~\ref{fig:overview}. Given context $h$, the policy $\pi_\theta$ generates reasoning followed by tool calls or a user reply. Interaction outcomes train task completion; token-level state contrasts train reasoning and actions to fit the current conditions.

\subsection{Constructing Paired Task States}
\label{sec:state-construction}

We construct data from ToolMind \citep{yang2025toolmind} and LoopTool-23k \citep{zhang2026looptool}. Structural checks remove invalid calls and inconsistent histories. We use Qwen3.8-27B to filter calls in ToolMind's synthetic demonstrations that do not serve the request or contain unsupported arguments. The same model rewrites user messages and synthesizes retrieval steps.

For each selected call, we identify its required tools and information, then modify the tool definitions or history to construct five types of states. \textsc{Exec} supplies the information and tools for execution. \textsc{Need\_User} removes a required user-provided value or retains a clarification turn, requiring the agent to ask before executing. \textsc{Need\_Fetch} removes a relevant tool result or introduces a lookup for a required identifier, so a tool can obtain the missing information. \textsc{Infeasible} removes the required tool and its identified alternatives. \textsc{Stop} includes a result completing the request and requires a reply without further calls.

Each non-\textsc{Exec} instance is paired with an \textsc{Exec} counterpart from the same source trajectory when available. Source calls and observations provide execution checks and interaction feedback; unpaired instances receive outcome supervision alone. Resampling by condition yields approximately 30k instances. Appendix~\ref{app:construction} details construction and annotation, with prompts in Appendix~\ref{app:prompts}.

\subsection{Learning from Interaction Outcomes}
\label{sec:interaction-outcomes}

An appropriate first action must lead to an appropriate follow-up action. Asking for information should enable execution once it arrives; receiving a result should lead to further work or termination according to the remaining task. We train short interactions in which clarification receives stored user information; retrieval receives recorded observations when function names and designated arguments match. The policy acts on the updated history. If a continuation is recorded, successful target execution is followed by tool results and one further policy step, checked against the requirement to continue or stop. For a trajectory $\tau$ requiring execution,
\begin{equation}
\label{eq:outcome-reward}
R_{\mathrm{out}}(\tau)=
\begin{cases}
\gamma^{k-1}\alpha(\tau)c(\tau), & \text{success},\\
0, & \text{otherwise}.
\end{cases}
\end{equation}
Success requires matching designated functions and arguments. In \textsc{Need\_User} and \textsc{Need\_Fetch}, attempting target execution before obtaining required information ends the trajectory with zero reward. Here, $k$ is the execution step, $\gamma=0.8$ favors timely completion, and $\alpha(\tau)$ penalizes extra calls. The indicator $c(\tau)$ requires a successful continuation, or equals one if none is specified. Thus, reward depends on both target execution and acting appropriately after the state changes. \textsc{Infeasible} and \textsc{Stop} use terminal checks for the required non-call response.

Groups sampled from each state define relative advantages for a token-normalized GRPO objective $\mathcal J_{\mathrm{out}}$ \citep{shao2024deepseekmath}, with KL regularization to the base model $\pi_0$. This objective trains all policy outputs, including continuations. Appendix~\ref{app:training} gives reward settings.

\subsection{Learning from State Contrasts}
\label{sec:state-contrasts}

Outcome supervision evaluates actions over an interaction. State contrasts additionally train how generated reasoning and actions fit the current conditions. For paired contexts $(h_i,\tilde h_i)$, reference paths $z_i,\tilde z_i$ describe appropriate actions and continuations. We construct them from source calls and responses, adding clarification or retrieval as required. Appendix~\ref{app:training} provides a scoring example. The base model $\pi_0$, fixed during training, scores each policy-generated token $y_{i,t}$ under both contexts using the same prefix $p_{i,t}$:
\begin{equation}
\label{eq:state-contrast}
\begin{aligned}
r_{i,t}=\operatorname{clip}\bigl(&
\log \pi_0(y_{i,t}\mid h_i,z_i,p_{i,t})\\
&-\log \pi_0(y_{i,t}\mid\tilde h_i,\tilde z_i,p_{i,t}),\,-1,1\bigr).
\end{aligned}
\end{equation}
For example, a missing-value state requires clarification before execution, whereas its counterpart permits immediate execution. Positive scores favor content fitting the current state and path; negative scores favor the counterpart. Content fitting both similarly receives near-zero scores. A response can acknowledge a missing value and later speculate that it is available. These parts can receive opposing preferences that a single outcome score cannot express. The prefix contains preceding policy outputs and returned feedback. Paired contexts and paths are used only for scoring; the policy acts from its tools and history.

The scores weight generated tokens in the state-contrastive objective:
\begin{equation}
\label{eq:pair-objective}
\begin{aligned}
\mathcal J_{\mathrm{pair}}(\theta)={}&
\frac{1}{Z}\sum_{(i,t)\in\mathcal P}\operatorname{sg}[r_{i,t}]\\
&\quad\cdot\log\pi_\theta(y_{i,t}\mid h_i,p_{i,t}).
\end{aligned}
\end{equation}
Here, $\mathcal P$ contains scored reasoning and action positions in paired instances before the appended continuation; $Z$ counts all policy-generated tokens before that step across the batch; $\operatorname{sg}$ stops gradients. We maximize
\begin{equation}
\label{eq:joint-objective}
\mathcal J(\theta)=\mathcal J_{\mathrm{out}}(\theta)
+\lambda\mathcal J_{\mathrm{pair}}(\theta).
\end{equation}
In all-failure or equally rewarded all-success groups, outcome advantages vanish, but reasoning and actions can still differ in state compatibility. The pair objective provides supervision beyond these tied outcomes. We prioritize groups with varying rewards and retain a portion of paired successful groups, as detailed in Appendix~\ref{app:training}. Paired all-failure groups also contribute when retained.

\begin{table*}[t]
\centering
\small
\setlength{\tabcolsep}{1.6pt}
\renewcommand{\arraystretch}{1.0}
\newcommand{\leaderscore}[1]{\makebox[1.9em][c]{#1}}
\newcommand{\leadergain}[2]{\makebox[5.0em][c]{\leaderscore{#1}%
  \rlap{\if\relax\detokenize{#2}\relax\else\textsubscript{\fontsize{6}{7}\selectfont\textcolor{black!65}{$#2$}}\fi}}}
\begin{tabular*}{\textwidth}{@{\extracolsep{\fill}}l*{9}{c}}
\toprule
\multirow{2}{*}{Model} & \multicolumn{5}{c}{\bfcl{} Multi-turn $\uparrow$} & \multicolumn{1}{c}{ToolSandbox $\uparrow$} & \multicolumn{1}{c}{When2Call $\uparrow$} & \multicolumn{2}{c}{SimpleToolHalluBench $\downarrow$} \\
\cmidrule(lr){2-6}\cmidrule(lr){7-7}\cmidrule(lr){8-8}\cmidrule(l){9-10}
 & \multicolumn{1}{c}{Overall} & \multicolumn{1}{c}{Base} & \multicolumn{1}{c}{MF} & \multicolumn{1}{c}{MP} & \multicolumn{1}{c}{LC} & \multicolumn{1}{c}{Similarity} & \multicolumn{1}{c}{Accuracy} & \multicolumn{1}{c}{NTA} & \multicolumn{1}{c}{DT} \\
\midrule
\multicolumn{10}{@{}l}{\textit{Base models}} \\
Qwen3-4B & \leadergain{39.8}{} & \leaderscore{46.6} & \leaderscore{43.8} & \leaderscore{34.6} & \leaderscore{34.3} & \leadergain{67.4}{} & \leadergain{67.9}{} & \leadergain{49.9}{} & \makebox[6.0em][c]{\leadergain{28.4}{}} \\
Qwen3-8B & \leadergain{45.5}{} & \leaderscore{50.3} & \leaderscore{49.1} & \leaderscore{37.9} & \leaderscore{44.6} & \leadergain{73.4}{} & \leadergain{66.4}{} & \leadergain{59.8}{} & \makebox[6.0em][c]{\leadergain{28.2}{}} \\
Qwen3-14B & \leadergain{47.6}{} & \leaderscore{56.4} & \leaderscore{48.3} & \leaderscore{37.2} & \leaderscore{48.4} & \leadergain{74.6}{} & \leadergain{70.6}{} & \leadergain{53.9}{} & \makebox[6.0em][c]{\leadergain{\textbf{20.7}}{}} \\
Granite-4.2-8B & \leadergain{49.4}{} & \leaderscore{62.7} & \leaderscore{40.0} & \leaderscore{43.3} & \leaderscore{51.7} & \leadergain{82.1}{} & \leadergain{75.6}{} & \leadergain{20.8}{} & \makebox[6.0em][c]{\leadergain{33.3}{}} \\
Llama-3.1-8B-Instruct & \leadergain{8.0}{} & \leaderscore{8.3} & \leaderscore{8.7} & \leaderscore{7.5} & \leaderscore{7.7} & \leadergain{35.6}{} & \leadergain{34.8}{} & \leadergain{72.6}{} & \makebox[6.0em][c]{\leadergain{98.8}{}} \\
\midrule
\multicolumn{10}{@{}l}{\textit{Baselines}} \\
GRPO-8B & \leadergain{51.9}{} & \leaderscore{58.7} & \leaderscore{54.1} & \leaderscore{45.4} & \leaderscore{49.4} & \leadergain{78.2}{} & \leadergain{71.7}{} & \leadergain{\textbf{55.6}}{} & \makebox[6.0em][c]{\leadergain{26.6}{}} \\
ToolRL-8B & \leadergain{48.9}{} & \leaderscore{55.1} & \leaderscore{51.3} & \leaderscore{39.1} & \leaderscore{50.3} & \leadergain{72.5}{} & \leadergain{65.1}{} & \leadergain{58.7}{} & \makebox[6.0em][c]{\leadergain{31.1}{}} \\
LoopTool-8B & \leadergain{53.1}{} & \leaderscore{61.2} & \leaderscore{53.7} & \leaderscore{43.8} & \leaderscore{53.8} & \leadergain{76.9}{} & \leadergain{68.8}{} & \leadergain{64.3}{} & \makebox[6.0em][c]{\leadergain{28.4}{}} \\
AWM-4B & \leadergain{41.2}{} & \leaderscore{50.0} & \leaderscore{43.0} & \leaderscore{34.4} & \leaderscore{37.3} & \leadergain{68.2}{} & \leadergain{68.8}{} & \leadergain{54.7}{} & \makebox[6.0em][c]{\leadergain{28.7}{}} \\
AWM-8B & \leadergain{46.6}{} & \leaderscore{53.5} & \leaderscore{49.2} & \leaderscore{38.1} & \leaderscore{45.8} & \leadergain{71.9}{} & \leadergain{60.9}{} & \leadergain{65.9}{} & \makebox[6.0em][c]{\leadergain{28.2}{}} \\
AWM-14B & \leadergain{\textbf{55.5}}{} & \leaderscore{66.0} & \leaderscore{52.3} & \leaderscore{\textbf{48.3}} & \leaderscore{55.4} & \leadergain{74.2}{} & \leadergain{70.2}{} & \leadergain{53.9}{} & \makebox[6.0em][c]{\leadergain{23.8}{}} \\
\midrule
\multicolumn{10}{@{}l}{\textit{\methodname{}}} \\
Qwen3-4B & \leadergain{\textbf{48.3}}{+8.5} & \leaderscore{\textbf{55.8}} & \leaderscore{\textbf{49.0}} & \leaderscore{\textbf{43.3}} & \leaderscore{\textbf{45.0}} & \leadergain{\textbf{73.3}}{+5.9} & \leadergain{\textbf{71.4}}{+3.6} & \leadergain{\textbf{44.1}}{-5.7} & \makebox[6.0em][c]{\leadergain{\textbf{24.3}}{-4.1}} \\
Qwen3-8B & \leadergain{\textbf{55.2}}{+9.8} & \leaderscore{\textbf{61.4}} & \leaderscore{\textbf{56.1}} & \leaderscore{\textbf{48.8}} & \leaderscore{\textbf{54.6}} & \leadergain{\textbf{79.3}}{+5.9} & \leadergain{\textbf{72.4}}{+6.0} & \leadergain{56.1}{-3.7} & \makebox[6.0em][c]{\leadergain{\textbf{25.0}}{-3.2}} \\
Qwen3-14B & \leadergain{54.8}{+7.3} & \leaderscore{\textbf{66.0}} & \leaderscore{\textbf{52.5}} & \leaderscore{45.4} & \leaderscore{\textbf{55.4}} & \leadergain{\textbf{81.3}}{+6.7} & \leadergain{\textbf{74.2}}{+3.7} & \leadergain{\textbf{45.0}}{-8.9} & \makebox[6.0em][c]{\leadergain{22.6}{+1.9}} \\
Granite-4.2-8B & \leadergain{\textbf{53.8}}{+4.4} & \leaderscore{\textbf{66.5}} & \leaderscore{\textbf{43.1}} & \leaderscore{\textbf{46.4}} & \leaderscore{\textbf{59.0}} & \leadergain{\textbf{82.3}}{+0.2} & \leadergain{\textbf{78.3}}{+2.8} & \leadergain{\textbf{10.5}}{-10.4} & \makebox[6.0em][c]{\leadergain{\textbf{26.4}}{-7.0}} \\
Llama-3.1-8B-Instruct & \leadergain{\textbf{26.1}}{+18.1} & \leaderscore{\textbf{34.3}} & \leaderscore{\textbf{17.3}} & \leaderscore{\textbf{21.7}} & \leaderscore{\textbf{31.2}} & \leadergain{\textbf{44.6}}{+8.9} & \leadergain{\textbf{59.6}}{+24.8} & \leadergain{\textbf{15.7}}{-57.0} & \makebox[6.0em][c]{\leadergain{\textbf{42.6}}{-56.2}} \\
\bottomrule
\end{tabular*}
\caption{Comparison of \methodname{} with base models and tool-use training baselines on task performance and tool hallucination across four benchmarks. ToolSandbox similarity reflects how well execution trajectories satisfy task milestones. MF, MP, and LC denote Missing Functions, Missing Parameters, and Long Context. NTA and DT denote the no-tool and distractor-tool settings. All scores are on a 0--100 scale. Gray subscripts indicate score differences from the corresponding base model; \textbf{bold} marks the best result for each base model and size.}
\label{tab:main}
\end{table*}

\section{Experiments}
\label{sec:experiments}

\subsection{Experimental Setup}
\label{sec:experimental-setup}

\paragraph{Base Models.}
We train Qwen3-4B/8B/14B \citep{yang2025qwen3}, Granite-4.2-8B \citep{ibm2026granite42}, and Llama-3.1-8B-Instruct \citep{dubey2024llama3} to evaluate \methodname{} across model sizes and families.

\paragraph{Baselines.}
Comparisons include ToolRL \citep{qian2025toolrl}, LoopTool \citep{zhang2026looptool}, AWM \citep{wang2026awm}, and the base models. ToolRL uses Qwen3-8B trained on RLLA-4k; LoopTool and AWM use released models. Each retains its own training data. Single-step binary GRPO on our constructed states compares training objectives.

\paragraph{Benchmarks.}
\bfcl{} \citep{patil2025bfcl} measures multi-turn tool-use accuracy. ToolSandbox \citep{lu2025toolsandbox} measures task progress through trajectory similarity. When2Call \citep{ross2025when2call} measures macro accuracy across calling, clarification, and infeasibility. SimpleToolHalluBench \citep{yin2026reasoning} measures hallucination with no available tool (NTA) or only an irrelevant tool (DT).

\paragraph{Implementation Details.}
We train with LoRA \citep{hu2022lora} on eight NVIDIA H100 80GB GPUs with $\lambda=0.1$. Qwen3 component comparisons use the step-400 checkpoint. Tables~\ref{tab:main} and~\ref{tab:ablations} report evaluation means, with BFCL standard deviations in Table~\ref{tab:ablations}. Appendices~\ref{app:training} and~\ref{app:evaluation} provide training and evaluation settings.

\subsection{Main Results}
\label{sec:main-results}

On Qwen3-8B, \methodname{} leads the compared methods on \bfcl{}, ToolSandbox, and When2Call in Table~\ref{tab:main}, including all four BFCL categories. Its largest advantage over LoopTool is in Missing Parameters, which requires obtaining information before execution. Higher Missing Functions and Long Context scores indicate better handling of tool availability and longer histories. Hallucination decreases in both NTA and DT relative to the Qwen3-8B base model and LoopTool.

Across all five base models, \methodname{} improves every BFCL category and When2Call accuracy and reduces NTA hallucination. On Qwen3, it exceeds AWM on ToolSandbox at all sizes and BFCL at 4B and 8B. At 14B, BFCL accuracy is 54.8\%, close to AWM's 55.5\%, with higher ToolSandbox and When2Call scores and lower hallucination in both settings. No-tool hallucination is 45.0\%, compared with 53.9\% for AWM-14B.

\subsection{Ablation Studies}
\label{sec:ablations}

\begin{table}[t]
\centering
\footnotesize
\setlength{\tabcolsep}{2.2pt}
\renewcommand{\arraystretch}{1.08}
\begin{tabular*}{\columnwidth}{@{\extracolsep{\fill}}llccccc@{}}
\toprule
 & & \bfcl{} & TS & W2C & NTA$\downarrow$ & DT$\downarrow$ \\
\midrule
\multicolumn{7}{@{}l}{\textit{(A) State construction}} \\
GRPO & Original & 47.2\,{\scriptsize$\pm$1.8} & 73.7 & 66.9 & 63.3 & 37.2 \\
 & Constructed & \textbf{51.9\,{\scriptsize$\pm$0.5}} & \textbf{78.2} & \textbf{71.7} & \textbf{55.6} & \textbf{26.6} \\
ToolRL & Original & 48.9\,{\scriptsize$\pm$0.6} & 72.5 & 65.1 & 58.7 & 31.1 \\
 & Constructed & \textbf{53.0\,{\scriptsize$\pm$0.6}} & \textbf{78.5} & \textbf{71.3} & \textbf{55.5} & \textbf{29.8} \\
\midrule
\multicolumn{7}{@{}l}{\textit{(B) Training configuration}} \\
4B & Outcome only & 46.9\,{\scriptsize$\pm$1.3} & \textbf{73.6} & 69.7 & 54.7 & 31.8 \\
 & \methodname{} & \textbf{48.3\,{\scriptsize$\pm$1.0}} & 73.3 & \textbf{71.4} & \textbf{44.1} & \textbf{24.3} \\
8B & Outcome only & 53.0\,{\scriptsize$\pm$0.9} & 76.2 & 71.0 & 62.2 & 30.3 \\
 & \methodname{} & \textbf{55.2\,{\scriptsize$\pm$1.3}} & \textbf{79.3} & \textbf{72.4} & \textbf{56.1} & \textbf{25.0} \\
14B & Outcome only & 54.3\,{\scriptsize$\pm$1.5} & 81.0 & 72.8 & 56.9 & 25.2 \\
 & \methodname{} & \textbf{54.8\,{\scriptsize$\pm$1.1}} & \textbf{81.3} & \textbf{74.2} & \textbf{45.0} & \textbf{22.6} \\
\bottomrule
\end{tabular*}
\caption{State construction and training ablations. TS: ToolSandbox; W2C: When2Call; NTA, DT: hallucination rates (lower is better). \textbf{(A)} Qwen3-8B trained on original data (ToolMind for GRPO; RLLA-4k for ToolRL) and constructed states. \textbf{(B)} Outcome-only training and complete \methodname{} on constructed states. Bold: better of each pair.}
\label{tab:ablations}
\end{table}

Task-state construction improves training under different reward designs. Table~\ref{tab:ablations}A compares binary GRPO and ToolRL rewards on original data and constructed states. Both achieve higher BFCL, ToolSandbox, and When2Call scores, with lower NTA and DT hallucination. Neither comparison uses the pair objective.

On the same constructed data, complete \methodname{} reduces NTA and DT hallucination across Qwen3-4B, 8B, and 14B compared with outcome-only training in Table~\ref{tab:ablations}B. At 14B, NTA falls from 56.9\% to 45.0\%, and DT from 25.2\% to 22.6\%. When2Call accuracy improves at all three sizes, indicating better decisions about calling, clarification, and infeasibility. BFCL accuracy also improves at all three sizes. Appendix~\ref{app:dynamics} reports training dynamics.

\section{Analysis}
\label{sec:analysis}

\begin{figure*}[t]
\centering
\includegraphics[width=\textwidth]{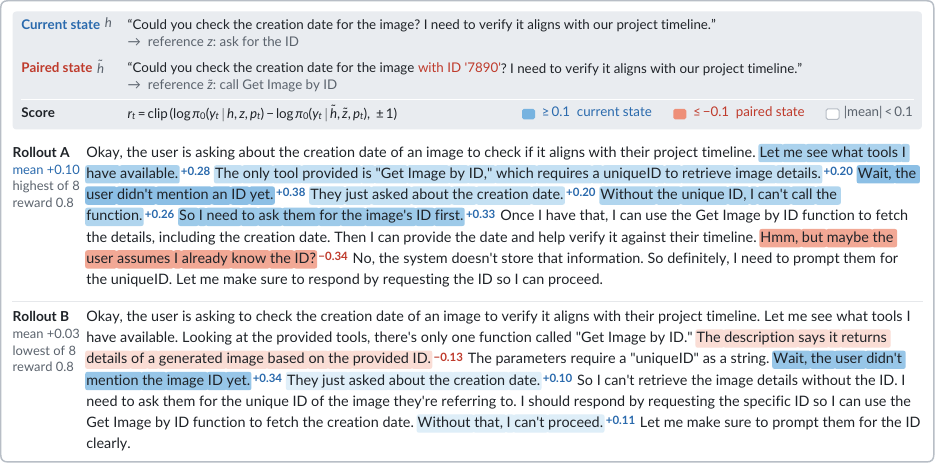}
\caption{State-contrast scores for two successful rollouts when the image ID is missing; the paired state provides it. Labels show mean token scores within each sentence. Blue indicates scores $\geq 0.1$, red indicates scores $\leq -0.1$, and unshaded sentences have scores between these thresholds. Both rollouts ask for the ID; their reasoning differs in state compatibility.}
\label{fig:state_contrast_rollout}
\end{figure*}

\subsection{State-Contrast Signals}
\label{sec:contrast-signals}

Successful rollouts can differ in how closely reasoning follows the available information. Figure~\ref{fig:state_contrast_rollout} pairs a request missing an image identifier with one that provides it. In rollout~A, recognizing the missing identifier and proposing clarification receive positive mean token scores; speculation that it is already known receives a negative score. Within one response, the signal rewards reasoning that fits the state and penalizes conflicting assumptions.

Both rollouts ask for the identifier and complete execution, but much of rollout~B's reasoning receives near-zero scores. All rollouts in this group share an outcome reward, giving zero relative outcome advantages. State contrasts still provide token-level supervision within this group. Appendix~\ref{app:contrast-signals} extends the analysis across all five conditions and describes example selection.

\subsection{From State Conditions to Actions}
\label{sec:state-action-consistency}

Table~\ref{tab:state_action}A revisits the diagnosis in Section~\ref{sec:task-state-decisions}. All four models identify the missing value in approximately 95\% of reasoning traces. Correct clarification among these responses reaches 83.2\% for \methodname{}, compared with 68.8\% for the base model and 80.6\% for LoopTool. Target-function selection when the value is available remains above 97\% for all four. \methodname{} improves clarification and retains accurate tool selection when execution is appropriate.

\begin{table}[!t]
\centering
\begingroup
\footnotesize
\setlength{\tabcolsep}{2pt}
\renewcommand{\arraystretch}{1.12}
\begin{tabular*}{\columnwidth}{@{\extracolsep{\fill}}lcccc@{}}
\toprule
\multicolumn{5}{@{}l}{\textit{(A) State-conditioned actions}} \\[2pt]
& \multicolumn{2}{c}{Value missing} & Value available & Paired states \\
\cmidrule(lr){2-3}\cmidrule(lr){4-4}\cmidrule(l){5-5}
Model & Mentions & Clarifies & Target call & Both correct \\
\midrule
Base & 94.9 & 68.8 & 97.7 & 63.5 \\
ToolRL & 94.6 & 70.4 & 97.9 & 65.1 \\
LoopTool & 95.4 & 80.6 & 98.0 & 75.4 \\
\methodname{} & 95.2 & 83.2 & 97.1 & 76.9 \\
\midrule
\multicolumn{5}{@{}l}{\textit{(B) Decisions when the value is missing}} \\[2pt]
Model & \multicolumn{2}{c}{Plans to execute $\downarrow$}
& \multicolumn{2}{c@{}}{Calls after deferring $\downarrow$} \\
\cmidrule(lr){2-3}\cmidrule(l){4-5}
Base & \multicolumn{2}{c}{9.9} & \multicolumn{2}{c@{}}{17.4} \\
ToolRL & \multicolumn{2}{c}{10.5} & \multicolumn{2}{c@{}}{16.4} \\
LoopTool & \multicolumn{2}{c}{10.2} & \multicolumn{2}{c@{}}{5.8} \\
\methodname{} & \multicolumn{2}{c}{6.4} & \multicolumn{2}{c@{}}{4.5} \\
\bottomrule
\end{tabular*}
\endgroup
\caption{State-conditioned reasoning and actions for Qwen3-8B (\%). \textbf{(A)} Mentions: identifies the missing value; Clarifies: correct clarification given that mention; Target call: correct function when the value is available; Both correct: mean product of per-state action accuracies within each pair. \textbf{(B)} Among traces identifying the missing value, Plans to execute measures proposals to call despite it; Calls after deferring measures target calls conditional on reasoning proposing to ask or take another step first. Lower is better for both metrics.}
\label{tab:state_action}
\end{table}

Panel~B distinguishes reasoning decisions from subsequent actions. Among traces identifying the missing value, \methodname{} proposes execution with an unprovided value or missing argument in 6.4\% of cases, compared with 10.5\% for ToolRL and 10.2\% for LoopTool. When reasoning proposes clarification or another step first, target calls still occur in 4.5\% of \methodname{} responses, compared with 5.8\% for LoopTool. Appendix~\ref{app:paired-diagnostics} covers other states and the irrelevant-history control.

\subsection{Multi-Step Execution and Recovery}
\label{sec:multi-step-analysis}

\begin{figure*}[t]
\centering
\includegraphics[width=\textwidth]{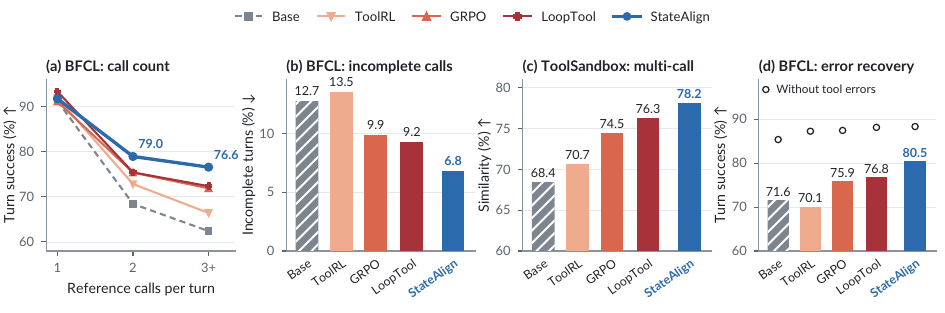}
\caption{Multi-step execution and recovery on Qwen3-8B. \textbf{(a)} \bfcl{} turn success by reference-call count. \textbf{(b)} Percentage of multi-call turns that fail after issuing at least one call but fewer than the reference requires. Panels a and b use ordinary execution turns reached without an earlier failure. \textbf{(c)} ToolSandbox similarity on multi-call scenarios. \textbf{(d)} \bfcl{} turn success with tool errors (bars) and without tool errors (open circles). Colors identify methods; arrows indicate the preferred direction.}
\label{fig:multicall_execution}
\end{figure*}

Multi-step tasks require agents to update what remains after each tool result and continue until the request is complete. Figure~\ref{fig:multicall_execution}a shows that \methodname{} leads turn success on two-call and three-or-more-call BFCL turns. Among its multi-call turns, 6.8\% fail after issuing some but fewer calls than required, compared with 9.9\% for GRPO and 9.2\% for LoopTool in panel~b. It also leads ToolSandbox multi-call similarity in panel~c. These results connect state-aware decisions to more complete execution of the required calls.

\begin{table}[t]
\centering
\footnotesize
\begin{tabular*}{\columnwidth}{@{\extracolsep{\fill}}lcc@{}}
\toprule
& BFCL & ToolSandbox \\
Model & Reply+fail $\downarrow$ & Similarity $\uparrow$ \\
\midrule
Base & 11.6 & 68.9 \\
GRPO & 5.7 & 71.3 \\
ToolRL & 11.4 & 69.3 \\
LoopTool & 6.3 & 67.7 \\
\methodname{} & \textbf{2.7} & \textbf{73.5} \\
\bottomrule
\end{tabular*}
\caption{Behavior and task completion after tool errors on Qwen3-8B. BFCL Reply+fail is the percentage of error-containing turns with no tool call in the next assistant response after the first error and a final failure. ToolSandbox reports similarity for error-containing scenarios. Values use a 0--100 scale; bold marks the best result in each column.}
\label{tab:error_recovery}
\end{table}

Tool errors identify conditions the agent must address before continuing. Figure~\ref{fig:multicall_execution}d shows a larger improvement over LoopTool on error-containing turns than on turns without errors. In Table~\ref{tab:error_recovery}, \methodname{} makes no call immediately after the first error and ultimately fails in 2.7\% of error-containing turns, compared with 5.7\% for GRPO and 6.3\% for LoopTool. \methodname{} also achieves the highest ToolSandbox similarity on error-containing scenarios.

A BFCL example illustrates these updates. After \texttt{startEngine} reports that the doors must be locked, the base model reports the failure and asks whether to lock them. \methodname{} locks the doors and retries. When the next error indicates that the brake pedal must be pressed, it performs that action and starts the engine successfully. Each result reveals a requirement for the next action, and the policy continues until the requested operation succeeds. Appendix~\ref{app:tool-error-recovery} provides the recorded sequence and further recovery comparisons.

\section{Related Work}
\label{sec:related-work}

\paragraph{Tool Learning.}
Synthetic datasets provide demonstrations of API calls, multi-turn interactions, and reasoning traces \citep{liu2025toolace,prabhakar2025apigenmt,yang2025toolmind}. Task construction extends this coverage to ambiguous, changing, and infeasible requests \citep{wang2026trajectory2task}. RL methods improve tool use through reward design \citep{zhang2026tooln1,qian2025toolrl}, adaptive data refinement \citep{zhang2026looptool}, and interactive training environments \citep{wang2026awm}. ARTIST trains interleaved reasoning and tool use \citep{singh2025artist}, and MUA-RL incorporates simulated users into policy rollouts \citep{zhao2025muarl}. Step-level credit assignment \citep{feng2025gigpo}, rewards for information acquisition \citep{kong2026infopo}, and self-distillation under privileged contexts \citep{zhao2026opsd} provide more fine-grained training feedback. SIPO contrasts contexts containing correct and incorrect answers \citep{yue2026sipo}; CC-OPD compares full instructions with contexts that omit individual constraints \citep{zheng2026ccopd}. RC-GRPO uses reward-conditioned sampling to increase reward diversity within rollout groups \citep{zhong2026rcgrpo}. \methodname{} contrasts task states that each have a valid action requirement, making their differences a source of supervision for how behavior should change.

\paragraph{Reliable Tool Use.}
Prior work combines reasoning with actions and observations \citep{yao2023react}, incorporates dialogue state tracking into tool-use training \citep{acikgoz2025coalm}, and separates state interpretation from action generation \citep{guo2026oodatool}. However, stronger reasoning can accompany increased tool hallucination \citep{yin2026reasoning}. TRUST addresses decision reliability through uncertainty-aligned rewards at key interaction turns \citep{zhou2026trust}. Proxy state-based evaluation uses inferred environment states to simulate interactions and assess task completion \citep{chuang2026proxystate}. We use paired states with different action requirements as supervision. \methodname{} compares the same generated reasoning and actions under both states and their reference paths to train behavior that fits each state.

\section{Conclusion}
\label{sec:conclusion}

We presented \methodname{}, combining task-state construction with state-contrastive reinforcement learning. Our diagnosis shows that agents can violate current information and tool conditions even after stating them in reasoning. Constructed states improve task performance and reliability under different reward designs. On the same data, the complete method further reduces tool hallucination compared with outcome-only training across three Qwen3 model sizes. Experiments across five base models show improved tool use, and behavioral analyses show more consistent decisions, more complete multi-call execution, and stronger recovery after tool errors.

\section*{Limitations}
\label{sec:limitations}

\methodname{} is trained on five types of constructed task states through short interactions with recorded feedback. Training feedback is limited to the interactions covered by the source trajectories. Extending this setup to longer workflows with changing user goals or tool behavior is a direction for future work. We evaluate five base models ranging from 4B to 14B parameters on text-based tool-use benchmarks. The effectiveness of the approach at larger scales and in multimodal tool environments remains to be studied.

\bibliography{references}
\appendix
\section{Training Data}
\label{app:construction}

\begin{table*}[t]
\centering
\small
\setlength{\tabcolsep}{4.5pt}
\begin{tabular*}{\textwidth}{@{\extracolsep{\fill}}llccccccc@{}}
\toprule
\textbf{Dataset} & \textbf{Size} & \textsc{Exec} & \textsc{Need\_Fetch} & \textsc{Need\_User} & \textsc{Infeasible} & \textsc{Stop} & Other & \textbf{Paired} \\
\midrule
ToolACE & 11,300 dialogues & 73.5 & 0.8 & 6.5 & 8.1 & 7.2 & 3.9 & --- \\
\href{https://huggingface.co/datasets/glaiveai/glaive-function-calling-v2}{Glaive-v2} & 112,960 dialogues & 23.5 & 0.0 & 5.0 & 8.6 & 24.0 & 38.9 & --- \\
APIGen-MT-5k & 5,000 dialogues & 19.7 & 27.6 & 26.5 & 0.4 & 12.2 & 13.6 & --- \\
When2Call & 15,000 samples & 0.0 & 0.0 & 49.6 & 50.3 & 0.0 & 0.1 & --- \\
ToolMind (synthetic) & 40,161 dialogues & 54.4 & 4.9 & 7.1 & 1.3 & 23.4 & 8.9 & --- \\
LoopTool-23k & 23,040 samples & 87.1 & 1.2 & 1.3 & 1.1 & 9.0 & 0.3 & --- \\
RLLA-4k & 4,000 samples & 86.9 & 0.7 & 2.9 & 3.2 & 3.8 & 2.5 & --- \\
\midrule
\textbf{StateAlign} & 30,433 samples & 58.7 & 9.9 & 13.1 & 6.5 & 11.8 & 0.0 & 55\% \\
\bottomrule
\end{tabular*}
\caption{Task-state composition of tool-use training datasets (\%). Size uses each release's reporting unit. Other covers reference responses outside the five task-state categories. Paired is the share of training samples linked to a counterpart requiring a different action; --- indicates that no such links are provided.}
\label{tab:dataset_comparison}
\end{table*}

\paragraph{Data construction.}
We apply structural checks and Qwen3.8-27B filtering to remove low-quality source examples. The same model rewrites user messages, synthesizes retrieval steps, and checks the resulting states. Missing-value rewrites preserve the request and other supplied information; validation removes cases where the omitted value remains inferable. Retrieval examples require a lookup before execution, and tool-availability checks account for alternative tools that could fulfill the request. Construction uses temperature 1.0, top-$p$ 0.95, and top-$k$ 20. Selected prompts are provided in Appendix~\ref{app:prompts}.

\paragraph{Training data.}
The training pool contains approximately 30k instances covering all five task states, with 55\% linked to a paired counterpart. Table~\ref{tab:dataset_comparison} reports their distribution alongside existing tool-use datasets.

\paragraph{Dataset comparison.}
We estimate each external dataset's distribution from 1,000 reference responses classified by Qwen3.8-27B using the tools, history, and reference response; \methodname{} uses its construction labels. Direct execution accounts for approximately 87\% of LoopTool-23k and RLLA-4k. When2Call mainly covers clarification and infeasibility, and retrieval and clarification are the largest categories in APIGen-MT-5k. Our data cover all five conditions and provide explicit state pairs.

The constructed collection is checked for question overlap with BFCL multi-turn and When2Call using normalized exact matching and n-gram comparisons. No question matches are detected under these checks.

\section{Training Details}
\label{app:training}

\begin{table*}[t]
\centering
\small
\begin{tabular*}{\textwidth}{@{\extracolsep{\fill}}lclc@{}}
\toprule
Hyperparameter & Value & Hyperparameter & Value \\
\midrule
Learning rate & $2\times10^{-5}$ & Trajectories per batch & 320 \\
KL coefficient & 0.04 & Policy clipping & 0.2 \\
LoRA rank & 64 & LoRA $\alpha$ & 128 \\
Trajectories per state & 8 & Rollout steps & 4 \\
Tokens per step & 1,536 & Sampling temperature & 1.0 \\
State-contrast coefficient & 0.1 & Successful-group share & 20\% \\
\bottomrule
\end{tabular*}
\caption{Training hyperparameters for \methodname{}. The successful-group share is the fraction of batch groups reserved for paired successful trajectories.}
\label{tab:training-hyperparameters}
\end{table*}

\paragraph{Training setup.}
We implement \methodname{} using TRL with LoRA adaptation and vLLM rollout generation. Training uses AdamW on eight NVIDIA H100 80GB GPUs, with the shared hyperparameters in Table~\ref{tab:training-hyperparameters}. The \methodname{} checkpoints are step 400 for Qwen3 and step 300 for Granite and Llama. For ToolRL on RLLA-4k, we use step 200.

\paragraph{Reward computation.}
Target execution is checked against the recorded function names and argument values. In Eq.~\ref{eq:outcome-reward}, extra calls set $\alpha(\tau)=0.5$; otherwise, $\alpha(\tau)=1$. The continuation check requires the recorded next execution or termination to succeed. Recorded user replies and tool observations are excluded from the policy loss. State-contrast scores are clipped to $[-1,1]$ and apply to reasoning and action tokens before the appended continuation; the continuation receives outcome supervision only.

\paragraph{State-contrast scoring.}
Both scoring inputs include the same instruction to identify the current goal, required information, and appropriate action. Each input also contains a reference path for its task state. These paths reuse the source calls and responses and specify clarification, retrieval, or non-execution when required; values obtained in later steps are described by their source rather than supplied in advance. Figure~\ref{fig:scoring-input} shows a recorded pair and the shared policy continuation. The additional instructions and reference paths are used only by the scorer.

\begin{figure*}[tp]
\centering
\fbox{\begin{minipage}{\dimexpr\textwidth-2\fboxsep-2\fboxrule\relax}
\small\raggedright
\textbf{Shared tool definition}\par
\texttt{Get Image by ID}: Returns the details of a generated image based on the provided `id`.\par
Required parameter: \texttt{uniqueID} (string), The unique ID of the generated image.
\par\medskip
\begin{minipage}[t]{0.48\linewidth}
\raggedright
\textbf{Current state $h$: identifier missing}\par
Could you check the creation date for the image? I need to verify it aligns with our project timeline.
\par\medskip
\textbf{Reference path $z$}\par
{\ttfamily 1. To do this I need the unique ID. Could you please provide it?\par
2. After the user replies: call `Get Image by ID` with the value from the reply.}
\end{minipage}\hfill
\begin{minipage}[t]{0.48\linewidth}
\raggedright
\textbf{Paired state $\tilde h$: identifier available}\par
Could you check the creation date for the image with ID '7890'? I need to verify it aligns with our project timeline.
\par\medskip
\textbf{Reference path $\tilde z$}\par
{\ttfamily 1. <tool\_call>\par
\{"name": "Get Image by ID",\allowbreak{} "arguments": \{"uniqueID": "7890"\}\}\par
</tool\_call>}
\end{minipage}
\par\medskip
\textbf{Policy-generated prefix $p$ shared by both scoring inputs}\par
{\ttfamily <think>
Okay, the user is asking to check the creation date of an image to verify it aligns with their project timeline. Let me see what tools I have available.\par
Looking at the provided tools, there's only one function called "Get Image by ID." The description says it returns details of a generated image based on the provided ID. The parameters require a "uniqueID" as a string. \par
Wait, the user didn't}
\par\medskip
\textbf{Next token scored under both inputs:} \texttt{\textvisiblespace mention}
\end{minipage}}
\caption{Excerpts of a recorded state-contrast scoring pair. The tool definition and policy-generated prefix are shared; the user message and reference path differ. The next token is scored under $(h,z,p)$ and $(\tilde h,\tilde z,p)$.}
\label{fig:scoring-input}
\end{figure*}

\paragraph{Ablation settings.}
The component ablations in Table~\ref{tab:ablations}B use the same constructed states, 320 trajectories per batch, and checkpoints at step 400 across Qwen3-4B, 8B, and 14B. Each configuration samples eight trajectories per starting state. The \methodname{} batch includes the reserved successful groups.

\section{Evaluation Settings}
\label{app:evaluation}

\begin{table*}[t]
\centering
\small
\begin{tabular*}{\textwidth}{@{\extracolsep{\fill}}lcccc@{}}
\toprule
Benchmark & Temperature & Top-$p$ & Top-$k$ & Response token limit \\
\midrule
\bfcl{} & 0.001 & 1.0 & --- & 4,096 \\
ToolSandbox & 0.6 & 0.95 & 20 & 16,384 \\
When2Call & 0.6 & 0.95 & 20 & Remaining context \\
SimpleToolHalluBench & 0.5 & 0.95 & 20 & 2,048 \\
\bottomrule
\end{tabular*}
\caption{Evaluation decoding settings for Qwen3-based models. The response limit includes reasoning tokens. A dash indicates no top-$k$ filtering; remaining context denotes the available token budget after the input.}
\label{tab:evaluation-settings}
\end{table*}

\paragraph{Decoding.}
BFCL and SimpleToolHalluBench use their benchmark-provided decoding defaults. When2Call retains each model's default generation configuration, and ToolSandbox uses a shared decoding configuration across model families. Table~\ref{tab:evaluation-settings} lists the resulting settings for Qwen3-based models. For BFCL, we apply YaRN \citep{peng2024yarn} to all Qwen3 models with scaling factor 2 and a 65,536-token serving context; other model families retain their native positional encodings. Each base model and its trained counterpart share the same evaluation settings and tool-call format.

\paragraph{Benchmarks.}
ToolSandbox uses 728 scenarios without RapidAPI dependencies and \texttt{deepseek-chat} as the user simulator. When2Call uses 300 examples balanced across tool calling, clarification, and infeasibility, and reports macro accuracy. SimpleToolHalluBench uses 296 examples for each of the no-tool and distractor-tool conditions. When2Call and SimpleToolHalluBench use their judge prompts with \texttt{deepseek-chat}.

\paragraph{Reported results.}
Tables~\ref{tab:main} and~\ref{tab:ablations} report means over six BFCL evaluations and three evaluations on each other benchmark. BFCL standard deviations in Table~\ref{tab:ablations} reflect repeated evaluations of the same checkpoint.

\section{Additional Analyses}
\label{app:diagnostics}

\subsection{State-Contrast Signals}
\label{app:contrast-signals}

The responses in Figure~\ref{fig:state_contrast_rollout} have the highest and lowest mean reasoning scores among eight rollouts sharing outcome reward 0.8; shading shows sentence means of token scores. We extend this comparison to 50 successful groups from steps 100--400, with ten groups per condition and eight rollouts per group. Mean first-response scores differ within every group despite identical outcome rewards, providing differentiated supervision when group-relative outcome advantages are zero.

\subsection{State-Conditioned Actions}
\label{app:paired-diagnostics}

The additional analyses retain the original Qwen3-8B outcome-only checkpoint. Table~\ref{tab:retrieval_diagnostic} extends the clarification analysis to 300 retrieval pairs from the training pool. Each pair requires a lookup when information is missing and execution when it is available. \methodname{} improves retrieval and paired action accuracy over Outcome only, retaining 97.0\% target-function accuracy. Removing irrelevant history in a separate 300-pair control preserves action types in 92.6--96.2\% of comparisons across Base, ToolRL, LoopTool, and \methodname{}.

\begin{table}[t]
\centering
\footnotesize
\begin{tabular*}{\columnwidth}{@{\extracolsep{\fill}}lccc@{}}
\toprule
Model & Retrieves & Executes & Both correct \\
\midrule
Base & 75.2 & 96.9 & 73.2 \\
ToolRL & 74.3 & 96.9 & 72.2 \\
LoopTool & 77.5 & 97.5 & 75.8 \\
Outcome only & 81.9 & \textbf{97.8} & 80.2 \\
\methodname{} & \textbf{88.2} & 97.0 & \textbf{85.8} \\
\bottomrule
\end{tabular*}
\caption{State-conditioned action accuracy on Qwen3-8B retrieval pairs (\%). Retrieves measures lookup selection when information must be fetched; Executes measures target-function selection when the required information is available. Both correct averages the product of the two per-state accuracies within each pair. Bold marks the best result in each column.}
\label{tab:retrieval_diagnostic}
\end{table}

Section~\ref{sec:task-state-decisions} uses 300 pairs per information condition and 276 for tool availability. The clarification pairs used in Section~\ref{sec:task-state-decisions} and Table~\ref{tab:state_action} are derived from source trajectories excluded from the training pool. Sampling uses four responses per state, temperature 0.6, top-$p$ 0.95, and a 4,096-token limit. Action checks evaluate response type and target-function selection. Figure~\ref{fig:task_state_diagnosis} counts unparsed labels as non-mentions; Mentions in Table~\ref{tab:state_action} averages per-pair rates over valid labels. Panel~B conditions on recognizing the missing value; Calls after deferring additionally requires an explicit decision to ask or take another step first. Decision judges see only reasoning.

\paragraph{Annotation validation.}
\label{app:judge-validation}
Qwen3.8-27B labels are compared with three independent Claude Opus annotations on 200 condition judgments and 80 decisions, using the same codebook and hiding source-model identities and original labels. Agreement with the majority label is 70.4\% for condition judgments and 90.0\% for decisions, excluding items without a majority. Rare labels are oversampled; population-weighted agreement is 97.2\% for condition judgments and 97.5\% for decisions. Additional trained-model checks yield 83.3\% agreement on 60 condition judgments and 100\% on 30 decisions.

\begin{figure*}[t]
\centering
\includegraphics[width=\textwidth]{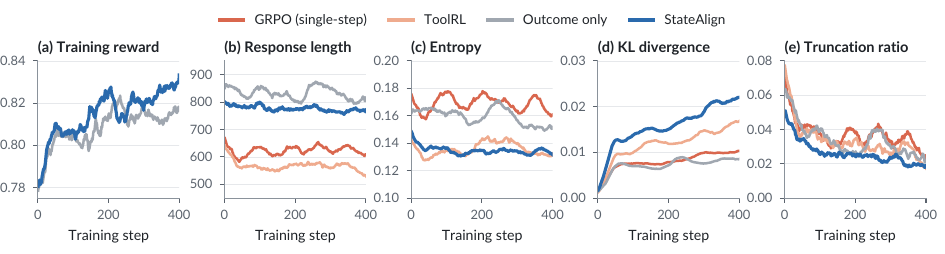}
\caption{Training dynamics on Qwen3-8B. (a) Mean outcome reward for Outcome only and \methodname{} on constructed states; (b) mean response length; (c) token entropy; (d) KL divergence to the base model; (e) the proportion of responses reaching the per-step token limit. Curves are 40-step centered moving averages.}
\label{fig:dynamics}
\end{figure*}

\subsection{Tool-Error Recovery}
\label{app:tool-error-recovery}

Table~\ref{tab:recovery_types} groups first-error messages by keywords, with precondition matches taking precedence. \methodname{} achieves the highest success rate in all three categories. Its advantage also persists at 333 shared case/turn positions where every method encounters an error in at least one evaluation; each policy contributes its own error-containing histories. This comparison supports the recovery result after restricting task positions.

\begin{table}[t]
\centering
\footnotesize
\begin{tabular*}{\columnwidth}{@{\extracolsep{\fill}}lcccc@{}}
\toprule
Model & Precondition & Not found & Other & Shared \\
\midrule
Base & 80.5 & 58.3 & 75.8 & 74.6 \\
GRPO & 84.8 & 62.2 & 77.7 & 77.4 \\
ToolRL & 79.6 & 56.1 & 76.4 & 72.8 \\
LoopTool & 89.9 & 63.5 & 69.5 & 79.3 \\
\methodname{} & \textbf{92.3} & \textbf{64.0} & \textbf{79.9} & \textbf{81.5} \\
\bottomrule
\end{tabular*}
\caption{BFCL recovery success on Qwen3-8B (\%). The first three columns group turns by the first tool error: an unmet precondition, a missing resource, or another error. Shared restricts the comparison to task positions where every method encounters a tool error. Bold marks the best result in each column.}
\label{tab:recovery_types}
\end{table}

Table~\ref{tab:recovery_sizes} extends recovery analysis to Qwen3-4B/8B/14B. Panel~A compares Outcome only with the complete \methodname{} configuration on BFCL. Panel~B compares base and trained policies on ToolSandbox, showing higher similarity after errors at all three sizes.

\begin{table}[t]
\centering
\footnotesize
\begin{tabular*}{\columnwidth}{@{\extracolsep{\fill}}lcc@{}}
\toprule
\multicolumn{3}{l}{\textit{(A) BFCL turn success}} \\
Base model & Outcome only & \methodname{} \\
\midrule
Qwen3-4B & 69.6 & \textbf{71.5} \\
Qwen3-8B & 78.4 & \textbf{80.5} \\
Qwen3-14B & 79.7 & \textbf{79.8} \\
\midrule
\multicolumn{3}{l}{\textit{(B) ToolSandbox similarity}} \\
Base model & Base & \methodname{} \\
\midrule
Qwen3-4B & 68.0 & \textbf{71.9} \\
Qwen3-8B & 68.9 & \textbf{73.5} \\
Qwen3-14B & 71.3 & \textbf{74.4} \\
\bottomrule
\end{tabular*}
\caption{Tool-error recovery across Qwen3 sizes. (A) BFCL turn success, comparing Outcome only with \methodname{}. (B) ToolSandbox similarity, comparing Base with \methodname{}. Both panels include only error-containing interactions. Higher is better on both 0--100 scales; bold marks the better result in each row.}
\label{tab:recovery_sizes}
\end{table}

\paragraph{Analysis settings.}
BFCL errors have an \texttt{error} field; ToolSandbox error messages begin with a named Error or Exception. Recovery pools official turn verdicts across six BFCL evaluations or similarity scores across three ToolSandbox evaluations. Eligible BFCL turns have no earlier conversation failure; ToolSandbox requires scored, readable trajectories. Reply+fail divides failed turns with no call immediately after the first error by all error-containing turns; absent responses are excluded from the numerator. Panels a--c of Figure~\ref{fig:multicall_execution} report mean per-evaluation scores. Panel~b counts failing turns with at least one but fewer than the required calls, divided by all eligible multi-call turns. Panel~d pools BFCL turns by error presence.

The case in Section~\ref{sec:multi-step-analysis} is BFCL \texttt{multi\_turn\_base\_93}, turn 2, evaluation 1. Its recovery sequence is \texttt{lockDoors}, \texttt{startEngine}, \texttt{pressBrakePedal}, and \texttt{startEngine}, ending with \texttt{engineState: running}.

\section{Training Dynamics}
\label{app:dynamics}

Figure~\ref{fig:dynamics} shows the first 400 training steps. GRPO uses binary single-step rewards on constructed states; ToolRL uses its reward on RLLA-4k. Outcome only and \methodname{} share the constructed states and outcome reward. Compared with Outcome only, \methodname{} reaches higher rewards late in training, with lower token entropy and greater KL divergence from the base model. It also generates shorter responses, and its truncation rate declines over training.

\section{Selected Prompts}
\label{app:prompts}

Figures~\ref{fig:prompt-source-filtering}--\ref{fig:prompt-retrieval-construction} present three prompts used with Qwen3.8-27B for source filtering and state construction. Each system prompt is paired with a user message containing the case-specific input described in its caption.

\begin{figure*}[tp]
\centering
\fbox{\begin{minipage}{\dimexpr\textwidth-2\fboxsep-2\fboxrule\relax}
\small\raggedright
\textbf{Source-data filtering}\par\medskip
\ttfamily
You check training data for a tool-calling assistant. Each case shows the tools the assistant can use, the conversation so far, and the assistant's next step (one or more tool calls). Judge that next step using only what the case shows, and answer six yes/no questions:

J1 user\_\allowbreak{}voice: Is the last user message written from the user's side (a request, or information the user provides), rather than in the assistant's voice (offering help, announcing an action, presenting results)?\par
J2 relevance: Does every call in the next step serve something the user asked for in this conversation that has not already been done?\par
J3 support: Is every argument value in the next step one of these: stated in the conversation; returned by an earlier tool result; a default or allowed value from the tool description; a standard conversion of such a value (units, formats, dates relative to today); common public knowledge (for example a company's ticker or a city's coordinates)? Answer no if any ID, account or order number, e-mail, URL, file name or personal detail appears to be made up.\par
J4 completeness: Does the next step pass every value the user gave for this request to the matching parameter, unchanged?\par
J5 better\_\allowbreak{}tool: Is there another tool in the list that clearly fits the user's request better than the tool(s) called?\par
J6 timing: At this point, should the assistant call a tool, rather than first ask the user for missing information or answer from what is already in the conversation?

Reply with one JSON object and nothing else:\par
\{"J1": "yes|no", "J2": "yes|no", "J3": "yes|no", "J4": "yes|no", "J5": "yes|no", "J6": "yes|no", "reason": "<one short sentence about every failing answer (J1-J4 or J6 no, J5 yes); empty if none>"\}

Example 1\par
TOOLS:\par
- get\_\allowbreak{}stock\_\allowbreak{}quote: Current price of a listed stock.\par
- symbol (string, required): Ticker symbol, e.g. MSFT\par
CONVERSATION:\par
USER: What is Tesla trading at right now?\par
NEXT STEP:\par
get\_\allowbreak{}stock\_\allowbreak{}quote(\{"symbol": "TSLA"\})\par
ANSWER: \{"J1": "yes", "J2": "yes", "J3": "yes", "J4": "yes", "J5": "no", "J6": "yes", "reason": ""\}

Example 2\par
TOOLS:\par
- get\_\allowbreak{}order\_\allowbreak{}status: Status of a customer order.\par
- order\_\allowbreak{}id (string, required): The order number.\par
- search\_\allowbreak{}products: Search the product catalog.\par
- query (string, required): Search words.\par
CONVERSATION:\par
USER: Where is my order? It was supposed to arrive yesterday.\par
NEXT STEP:\par
get\_\allowbreak{}order\_\allowbreak{}status(\{"order\_\allowbreak{}id": "ORD-12345"\})\par
ANSWER: \{"J1": "yes", "J2": "yes", "J3": "no", "J4": "yes", "J5": "no", "J6": "no", "reason": "The order number was never given; the assistant should ask for it."\}
\end{minipage}}
\caption{System prompt for filtering tool-use demonstrations. The input contains the available tools, conversation, and recorded next calls.}
\label{fig:prompt-source-filtering}
\end{figure*}

\begin{figure*}[tp]
\centering
\fbox{\begin{minipage}{\dimexpr\textwidth-2\fboxsep-2\fboxrule\relax}
\small\raggedright
\textbf{Missing-information construction}\par\medskip
\ttfamily
You edit user messages for a tool-calling dataset. You get the last user message of a conversation, the tool the assistant is about to call, one of its parameters, and the value the user stated for it. Rewrite the message so that the user no longer gives that value, and keep everything else.

Rules:\par
- Remove the value and anything that would let the assistant work it out: the same value written differently, an alias, a code, or a name that identifies it. If it appears more than once, remove every mention.\par
- Keep the request, every other detail (names, numbers, dates, preferences), the tone and roughly the length. Keep all other values exactly as written.\par
- The result must read naturally, as if the user simply forgot to mention it. Do not add placeholders such as [city] or "X", do not ask about the missing value, and do not say that something is missing.

Reply with one JSON object and nothing else: \{"rewritten": "<the rewritten user message>"\}

Example 1\par
MESSAGE: Can you check tomorrow's weather in Lyon for my cycling trip?\par
TOOL: get\_\allowbreak{}forecast | PARAMETER: city (Name of the city) | VALUE: "Lyon"\par
ANSWER: \{"rewritten": "Can you check tomorrow's weather for my cycling trip?"\}

Example 2\par
MESSAGE: Pull the latest SEC filings for Apple (AAPL), I only need the 10-K from 2023.\par
TOOL: get\_\allowbreak{}sec\_\allowbreak{}filings | PARAMETER: symbol (Stock ticker) | VALUE: "AAPL"\par
ANSWER: \{"rewritten": "Pull the latest SEC filings for the company I'm researching, I only need the 10-K from 2023."\}
\end{minipage}}
\caption{System prompt for constructing missing-information states. The input specifies the user message, target function, parameter description, and value to remove.}
\label{fig:prompt-missing-information}
\end{figure*}

\begin{figure*}[tp]
\centering
\fbox{\begin{minipage}{\dimexpr\textwidth-2\fboxsep-2\fboxrule\relax}
\small\raggedright
\textbf{Retrieval-state construction}\par\medskip
\ttfamily
You convert examples for training tool-calling assistants. In the example, the user gives an identifier (an id, code, username, slug, ...) and the assistant passes it to a function. Turn it into an example where the user describes the entity instead, so the assistant must first look the identifier up with a tool.

Produce:\par
1. rewritten: the user's message rewritten so that the user refers to the same entity by a natural description (its name, title, owner, date, place, or other attributes a user would know) and no longer gives the identifier or anything it can be derived from. Keep the request and every other detail, the tone and roughly the length.\par
2. lookup\_\allowbreak{}tool: a new read-only function \{name, description, parameters (JSON schema with properties and required)\} that searches such entities by those attributes and returns matching records with their identifiers. Use a name that is not already in the tool list.\par
3. lookup\_\allowbreak{}call: \{name, arguments\} - the call the assistant would make; every argument value must be copied verbatim from the rewritten message.\par
4. result: the JSON the lookup returns for that call: a list of 2-4 records with plausible fields; exactly one record matches the description and carries the identifier exactly as given; the others have different identifiers.

Reply with one JSON object and nothing else: \{"rewritten": "...", "lookup\_\allowbreak{}tool": \{...\}, "lookup\_\allowbreak{}call": \{...\}, "result": [...]\}

Example\par
USER MESSAGE: Can you pull the full order details for order ORD-55821?\par
NEXT CALL: get\_\allowbreak{}order\_\allowbreak{}details(\{"order\_\allowbreak{}id": "ORD-55821"\})\par
IDENTIFIER: order\_\allowbreak{}id = "ORD-55821"\par
ANSWER: \{"rewritten": "Can you pull the full order details for the blue standing desk I ordered on March 3rd?", "lookup\_\allowbreak{}tool": \{"name": "search\_\allowbreak{}orders", "description": "Search the customer's orders by product name and order date; returns order ids with summaries.", "parameters": \{"type": "object", "properties": \{"product": \{"type": "string", "description": "Product name or keywords"\}, "order\_\allowbreak{}date": \{"type": "string", "description": "Order date as written by the customer"\}\}, "required": ["product"]\}\}, "lookup\_\allowbreak{}call": \{"name": "search\_\allowbreak{}orders", "arguments": \{"product": "blue standing desk", "order\_\allowbreak{}date": "March 3rd"\}\}, "result": [\{"order\_\allowbreak{}id": "ORD-55821", "product": "Blue standing desk", "order\_\allowbreak{}date": "2024-03-03", "status": "shipped"\}, \{"order\_\allowbreak{}id": "ORD-55377", "product": "Blue desk lamp", "order\_\allowbreak{}date": "2024-02-19", "status": "delivered"\}]\}
\end{minipage}}
\caption{System prompt for constructing retrieval states. The input supplies the existing tool names, user message, target call, and identifier with its parameter description.}
\label{fig:prompt-retrieval-construction}
\end{figure*}

\end{document}